\documentclass[twocolumn]{svjour3}          % twocolumn
\usepackage[left=2cm,top=3cm, bottom=3cm, right=2cm]{geometry}

\usepackage{graphicx}
\begin{document}

\title{Efficient Supervision for Robot Learning 
\thanks{The author would like to acknowledge the support of the UK's Engineering and Physical Sciences Research Council (EPSRC) through the Doctoral Training Award (DTA), the Hans-Lenze-Foundation, the Dr-Jost-Henkel-Foundation, New College Oxford and the Department of Engineering Science.}
}
\subtitle{via Imitation, Simulation, and Adaptation}


\author{Dissertation Summary \\
Markus Wulfmeier
}


\institute{University of Oxford \at
           Oxford, United Kingdom \\
            %   Tel.: +123-45-678910\\
            %   Fax: +123-45-678910\\
              \email{m.wulfmeier@gmail.com}             \\
            \emph{Present position:} DeepMind, London, United Kingdom  
            \\         Pre-Print Version. The final publication is available at https://link.springer.com/article/10.1007/s13218-019-00587-0}
\maketitle

\begin{abstract}
\textit{Recent successes in machine learning have led to a shift in the design of autonomous systems, improving performance on existing tasks and rendering new applications possible. 
Data-focused approaches gain relevance across diverse, intricate applications when developing data collection and curation pipelines becomes more effective than manual behaviour design. 
The following work aims at increasing the efficiency of this pipeline in two principal ways: by utilising more powerful sources of informative data and by extracting additional information from existing data. 
In particular, we target three orthogonal fronts: imitation learning, domain adaptation, and transfer from simulation. 
}
\keywords{Data Efficiency \and Transfer Learning \and Inverse Reinforcement Learning \and Domain Adaptation \and Sim2Real}
\end{abstract}

\section{Introduction}

In order to enable more widespread application of robots when addressing complicated, repetitive, and burdensome tasks, we are required to minimise the  effort connected to the introduction of existing platforms to new environments and functions. Recent advances in machine learning have had considerable impact in this context, improving performance and generality of existing systems and rendering new tasks possible \cite{mnih2013playing,poplin2018prediction,silver2017mastering,urmson2008autonomous}.

Significant benefits are frequently obtained on the perception side of the software stack. In addition, other modules including prediction, mapping, planning, and control become increasingly affected by the potential gains of integrating machine learning.

\begin{figure}
	\centering
		\includegraphics[width = .32 \textwidth]{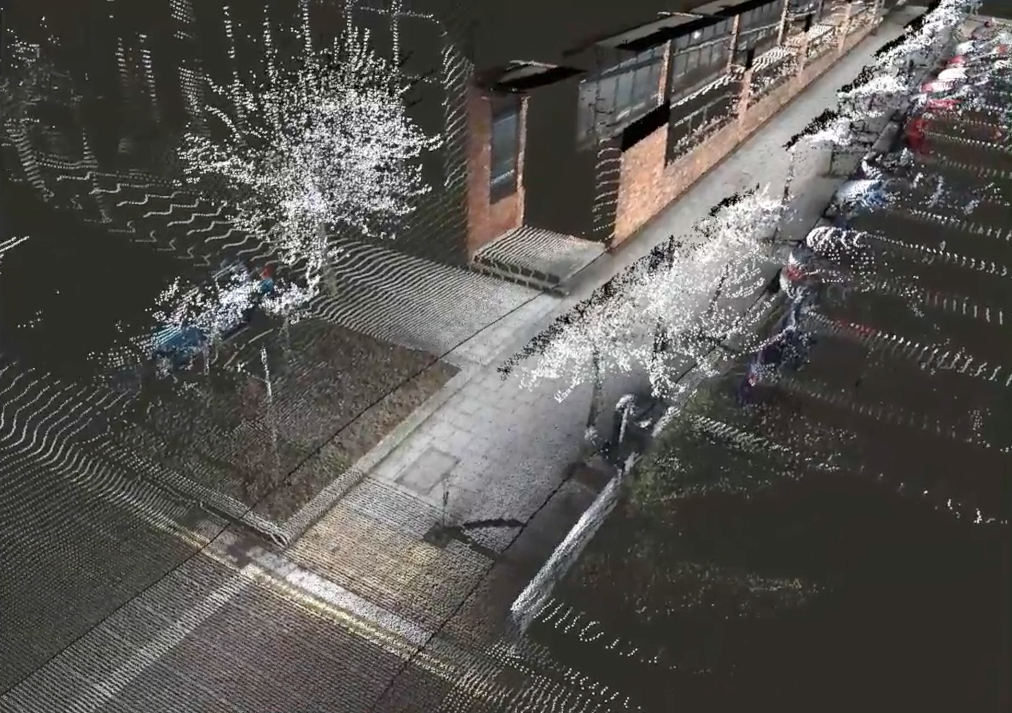}
    \caption{\small Coloured 3D LIDAR scan of a sidewalk representing a common part of the perception of autonomous platforms. The colour for the points is provided via side-facing cameras with known calibration to the LIDAR system.}
	\label{fig:mk}
\end{figure}

In the context of changes in the design and development of autonomous systems, human effort shifts strongly towards data collection, curation, and evaluation. In addition to ascribing a more dominant role to data, this process increases the need for thorough and exact tools for investigating and testing the impact that different sources and aggregations of data have on metrics of interest for the final system. 

We focus in this work on identifying efficient ways to generate informative data with respect to the task at hand as well as methods for optimising the utility of existing data. Essentially, the generation of unsupervised data, e.g. raw images without additional, structured information such as classification or segmentation of elements in the raw data, can be performed at low cost. Continuing with the example of images, additionally providing supervision such as pixel-wise segmentation labels presents a very labour-intensive process. 
However, not all additional information is generated at equal cost and the following sections aim to explore different strategies, including the use of imitation learning, domain adaptation, and transfer learning from simulation, in order to simplify the procedure and reduce human effort.

Given this perspective, imitation learning \cite{argall2009survey} represents a straightforward way for personnel without explicit robotics or programming background to teach robots to perform tasks by providing inexpensive demonstrations. We develop a scalable approach \cite{WulfmeierIROS2016,Wulfmeier2017IJRR} to identify the preferences underlying existing demonstrations via the framework of inverse reinforcement learning and enable the integration as cost maps into existing motion planning systems. 

In addition to employing low-cost labels from demonstrations, we investigate the adaptation of models to domains without available label information. 
Specifically, the challenge of appearance changes in outdoor robotics such as illumination and weather shifts is addressed using adversarial domain adaptation \cite{Ganin2016,wulfmeier2017addressing}. We demonstrate performance benefits of the method for semantic segmentation of drivable terrain and extend our approach to benefit from the continuity of changes in most outdoor domains \cite{wulfmeier2018incremental}. 

Finally, we investigate simulations as a common domain for the generation of immense amounts of data at comparably low cost. The approach focuses on transfer learning \cite{wulfmeier17matl} in situations, where the characteristic differences extend past visual appearance to the underlying system dynamics. From this perspective, our work aims at parallel training in both systems and mutual guidance via auxiliary alignment objectives to accelerate training for real world systems. 

We demonstrate increased performance and data efficiency across all projects, rendering diverse implementations of robotic platforms closer to required accuracy and safety metrics for large-scale application. We conclude by indicating current shortcomings of the presented work and directions for future improvements. For a broad survey of the related literature, the interested reader is referred to the complete thesis \cite{wulfmeier2018efficient}.

\section{Imitation}

The field of learning from demonstration (LfD) provides a direct way to model
robot behaviours without excessive knowledge about programming or robotics. 
Building on task demonstrations enables us to benefit from existing expert knowledge in new domains. While other approaches to providing supervision result in potentially large additional amounts of manual annotation (e.g. in pixel-wise segmentation), LfD can enable us to efficiently generate large amounts of supervised data.

\begin{figure}
	\centering
		\includegraphics[width = .45 \textwidth]{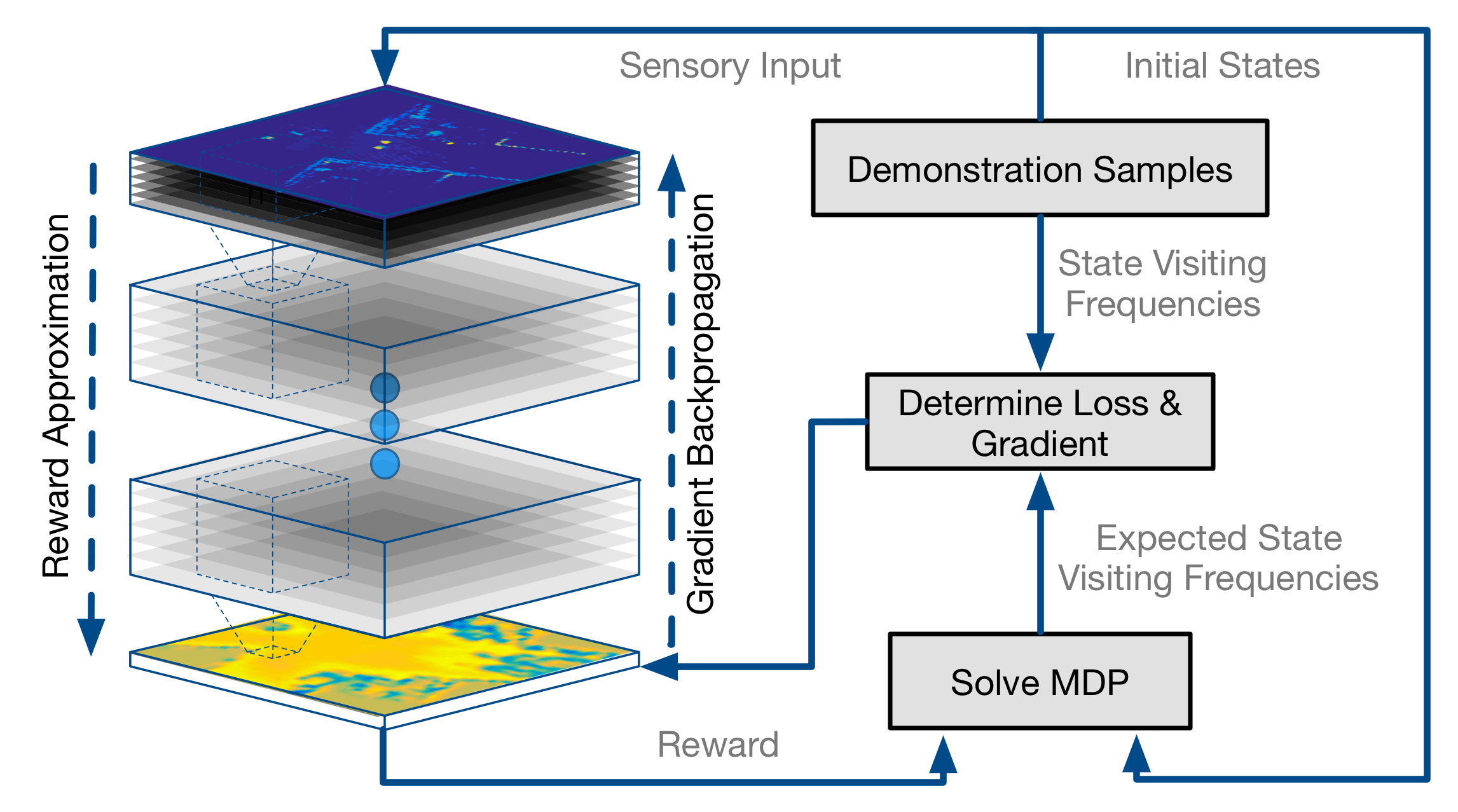}
    \caption{\small Schema for training deep neural networks in the Maximum Entropy paradigm for IRL. A randomly initialised network mapping from observations to reward is iteratively refined to align an reinforcement learning agent's policy to the respective demonstration trajectories.}
	\label{fig:medirl}
\end{figure}

The two principal ways to represent extracted behaviour are given by policies $\pi$ and reward functions $r$. While the former directly encodes behaviour as a state $s$ to action $a$ mapping $\pi(a|s)$, the latter more generally describes preferences for parts of the state and action spaces as $r(s,a)$. The complete system is commonly represented as a Markov Decision Process (MDP), which is described in more detail in Chapter 2 of  \cite{wulfmeier2018efficient}. 

As a succinct and flexible representation, with straightforward incorporation into existing modular software pipelines as well as easier visualisation and interpretation, we focus on the reward function to encode behaviour for application in the context of autonomous mobility. Essentially, inverse reinforcement learning (IRL) methods simulate the behaviour of a reinforcement learning (RL) agent and optimise the reward function such that agent behaviour and demonstration data aligns (as schematised in Figure \ref{fig:medirl}).

Prior work \cite{ziebart2008maximum,levine2011nonlinear} enables the representation of complex behaviours given well-structured, manually crafted input representations. However, in order to improve scalability with respect to complexity of the environment when presented with raw data (as exemplified in Figure \ref{fig:mk}), we develop Maximum Entropy Deep Inverse Reinforcement Learning (MEDIRL). The method extends previous methods for linear function approximators \cite{ziebart2008maximum} to more flexible deep function approximators conducive to address data sources with higher variability and dimensionality as well as greater amounts of training data. 

Building on fully convolutional deep neural networks, we present a method which is able to learn from large datasets of human driving to extract the underlying preferences and utilise the extracted knowledge as cost maps for motion planning. 
The approach reduces the requirements for manual feature and model design and facilitates the use of raw data to learn complex representations.

Following first experiments on common toy datasets, the approach is evaluated on a large 120 km driving dataset based on 3D LIDAR scans. The data were collected over the course of a year with 13 different drivers, resulting in a total of 25.000 trajectory samples and enables our method to outperform manually crafted cost functions with respect to the classification of traversable paths and prediction of human behaviour. In addition, successor works integrate human intuition and domain knowledge for cost function design and extends the method to incorporate manually designed cost functions into a model pretraining step. 
Finally, we demonstrate robustness towards systematic noise in form of miscalibration of the sensor setup.

Shortcomings of the current instantiation of the approach include the limited applicability to more dynamic environments as the planning or reinforcement learning step in the current method instantiation only considers the sensor input of the first step along the trajectory. 
To address the partial observability introduced by this procedure, future work can build on the incorporation of sequences of input data as well as sequence predictions for the cost function. Both of the directions are increasingly rendered more computationally tractable due to improvements in state-of-the-art graphics processing units and other more specialised processors.

\section{Adaptation}
In addition to utilising more efficient sources of data, such as human demonstrations, we aim to maximise the utility of available data. This direction becomes particularly relevant for perception systems which, unlike LIDAR, are significantly affected by  changing conditions such as weather and daylight during deployment. Camera-based setups present one example of this class, where models trained under specific conditions commonly fail to generalise to others without sufficiently diverse training data. 
However, as low-cost and high-resolution alternative these setups are predominant across a wide range of applications and deserve additional focus.

The long-standing challenge of reduced model performance based on appearance change during deployment is frequently addressed in the context of unsupervised domain adaptation. Current state-of-the-art results for transfer with common, smaller datasets such as MNIST and SVHN are obtained without additional constraints via adversarial domain adaptation (ADA, which is also known under domain adversarial networks) \cite{Ganin2016,bousmalis2016domain}. 

The contributions in the corresponding publications aim at the alignment between data representations between domains with and without supervision in order to improve model accuracy regarding the latter.
We investigate relevant aspects in the context of scaling applications of ADA to state-of-the-art architectures and streamlining domain adaptation for classification as well as semantic segmentation applications in outdoor robotics. 

Essentially, adversarial training schemes such as Generative Adversarial Networks (GAN) \cite{goodfellow2014generative} or ADA \cite{Ganin2016} are known to be notoriously unstable. 
Therefore, we begin our work by performing an extensive ablation study to identify relevant hyperparameters and further training details for the domain of interest, adapting techniques known to stabilise GAN training, and evaluate on a surrogate task with the same underlying appearance changes. The final application of our work lies in semantic segmentation of drivable terrain in an urban scenario, where we demonstrate a significant performance increase in the target domain by applying ADA over a FCN-VGG16 baseline trained only on source domain data.

\begin{figure}
	\centering
		\includegraphics[width = .5 \textwidth]{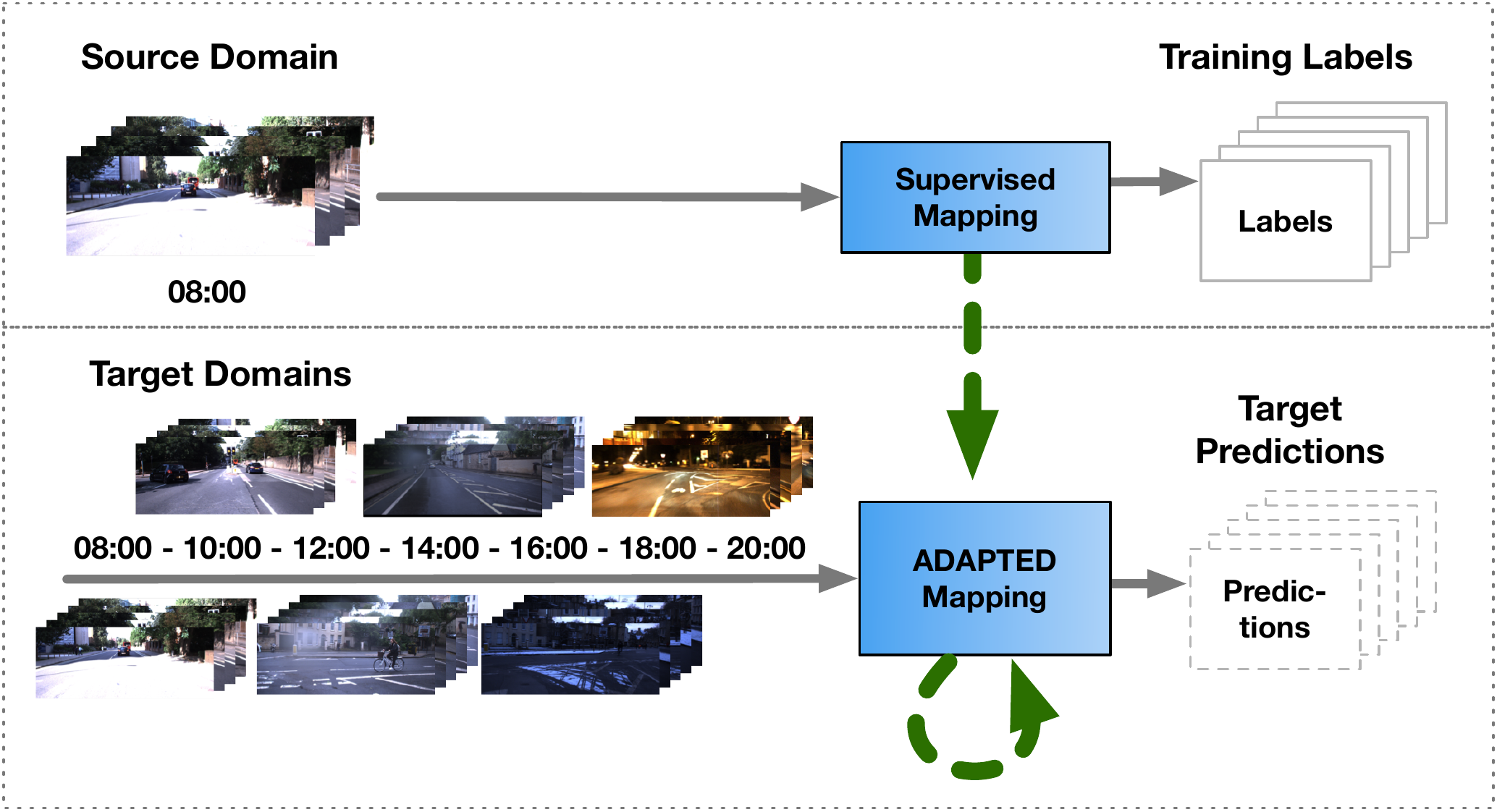}
    \caption{\small Incremental Adversarial Domain Adaptation (IADA). Instead of performing domain adaptation over larger appearance shifts at once, IADA splits domain alignment into simpler sub-tasks. After adapting the feature embedding of the initial target domain, the approach incrementally refines all modules to the currently perceived target domains.}
	\label{fig:iada}
\end{figure}

The method is extended in later work to exploit the incremental nature underlying most appearance shifts in outdoor domains as displayed in Figure \ref{fig:iada}. Furthermore, a generator model is introduced to imitate source domain data and to remove the requirement of storing immense amounts of the corresponding data modality \cite{wulfmeier2018incremental}, providing additional benefits for smaller, embedded systems with limited storage capacity.

\section{Simulation}
One domain of increasing influence is represented by simulations. Similarly to learning from demonstration, transfer learning from simulation builds on inexpensive data sources. Extensive amounts of data can be generated resulting only in the cost of increased computation and simulations consistently have gained in accuracy while the required resources become cheaper. 

However, simulations represent just an approximation of the real platform and models trained in simulation often fail to generalise when applied on the real robot. 
While the algorithms depicted in the previous section address differences in the observations between real systems and simulations, this section assumes equivalent state spaces to enable focus on varying system dynamics, including changes in friction, dampening and density. 

Simulations have particular relevance in the context of reinforcement learning (RL), where - in comparison to supervised and unsupervised learning - no dataset is given in advance but the agent collects its own data by observing and acting in its environment. 
The agent's behaviour is optimised in order to maximise received rewards which are provided from its environment. Simulations can provide extensive data sources, while at the same time creating the challenge of adapting learned modules to the real platform. 

\begin{figure}
	\centering
		\includegraphics[width = .37 \textwidth]{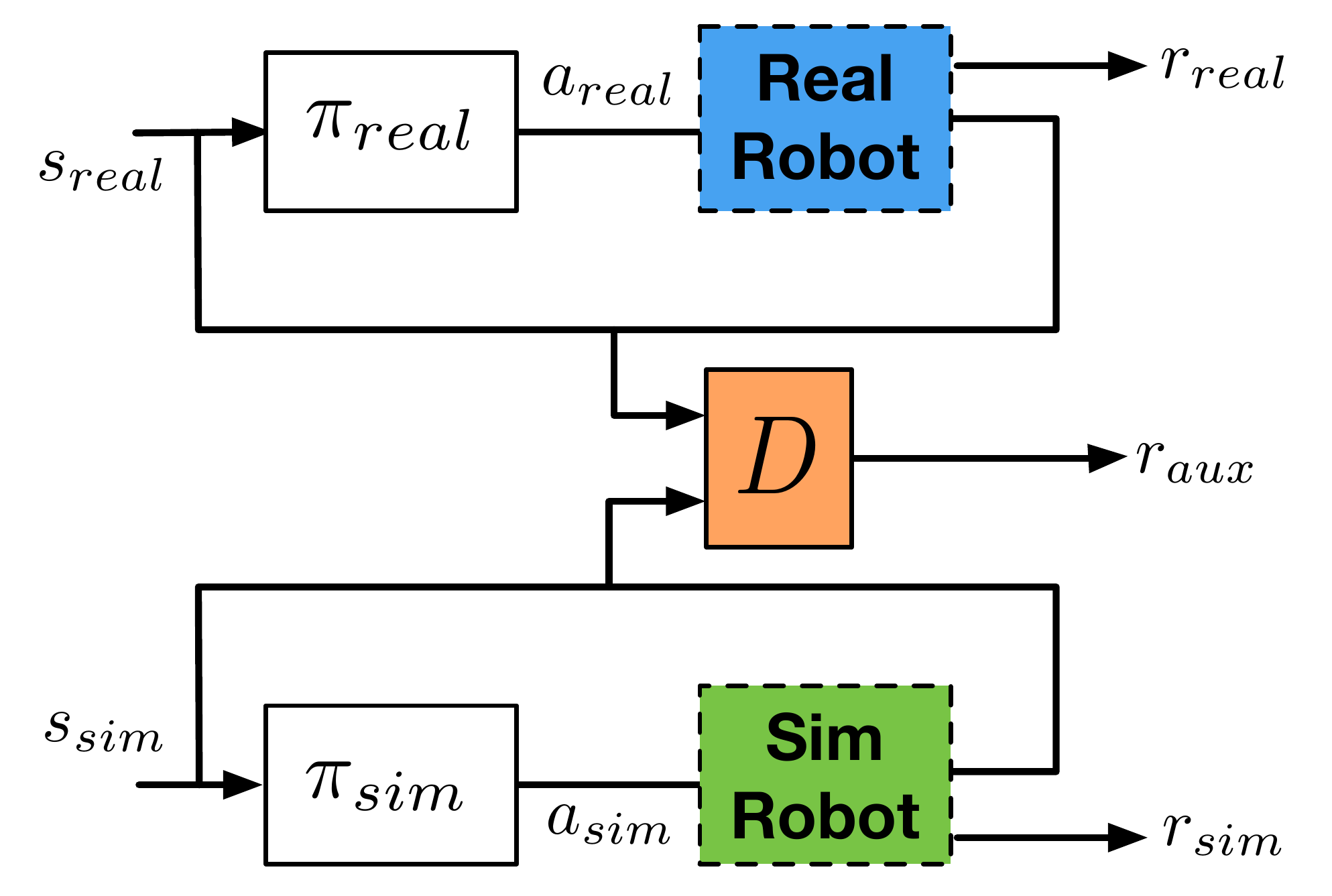}
    \caption{\small Simplified schema for Mutual Alignment Transfer Learning. Both policies $\pi_{sim/real}$ are trained to not only maximise their respective environment rewards, respectively $r_{sim/real}$, but also auxiliary alignment rewards $r_{aux}$ based on the discriminator $D$ which encourages both systems to occupy similar distributions over visited states. }
	\label{fig:matl}
\end{figure}

Mutual Alignment Transfer Learning (MATL) \cite{wulfmeier17matl} exemplifies a category of methods for parallel training in source and target domain which can guide exploration continuously on both systems while utilising information from each domain to further advance training on the other. 
For the purpose of mutually reinforcing training dynamics, we introduce auxiliary rewards into both systems aimed at the alignment of their respective distributions over visited states. 
The approach is conceptually related to generative adversarial imitation learning \cite{ho2016generative,stadie2017} where an agent's policy is optimised for one-sided alignment towards a set of demonstration trajectories. 

We extend existing work by incorporating a density estimator for both systems' distributions over visited states building on GANs \cite{goodfellow2014generative} and train the discriminator to classify the domain of origin for visited states or state sequences.  The auxiliary rewards are computed based on the inaccuracy of this discriminator as displayed in Figure \ref{fig:matl}.

In additional to improving performance over our baselines, the experiments demonstrate that mutual alignment improves over simple unilateral alignment of the robot towards the simulator agent's distribution over visited states. We hypothesise that the benefits rely on the agent in simulation being driven to explore better behaviour particularly for states visited by the robot agent, thus focusing on relevant parts of the state-space. 

When combining the presented algorithm with fine-tuning of policies pretrained in simulation, we can further improve performance in the majority of performed experiments. However, we lastly demonstrate that MATL succeeds even in transfer scenarios with significant changes in the dynamics when pretrained networks result in performance-degrading initialisation.

As MATL applies partially trained policies on the real robot we depend on additional safety measures such as virtual workspace constraints and torque limits in order to enable broader applicability in future work.

\section{Conclusion}
To enable the integration of autonomous platforms into real-world scenarios, we are required to address the immense complexity and variability of real environments as well as the modelling of intricate behaviours of interacting agents.

Machine learning and particularly deep learning have provided strong tools to address this challenge conditionally on the availability of large amounts of informative, labelled training data. 
While machine learning has accelerated progress across a broad range of applications including autonomous driving, logistics, resource allocation and many others, it has shifted efforts from model design to data handling and progress in many applications is still limited given the high cost and time requirements of data collection and annotation.

To improve the efficiency of the data pipeline and therefore commercial viability of robotic systems, we investigated approaches to reduce the cost of acquiring informative data as well as to maximise the utility of existing data. 
In particular, the three complementary methods presented here build on imitation learning, domain adaptation and transfer learning from simulation in order to reduce the effort connected to introducing robots into new environments and tasks. 

Acting in unstructured real-world environments and interacting with intricate agents while building on incomplete observations is a tremendously demanding challenge. 
Above all presented performance gains with individual approaches, we aim to provide support for a broader statement on the importance of transfer. 
Considering the long-term goal of extending the capabilities of our systems and improving generalisation, robustness and safety, 
we immensely benefit from, and lastly depend on, approaching correlated tasks, domains and sources of information jointly in order to build symbiotic, scalable solutions.

% \newpage
\linespread{1.0}
% \section{Author Vita}
% Markus Wulfmeier is research scientist at DeepMind focusing on efficient machine learning for robotics via transfer, decomposition and modularity with a strong background in simulation-to-real transfer, domain adaptation and learning from demonstration.

% Recently, he has been a postdoctoral research scientist at the Oxford Robotics Institute and a visiting scholar with the UC Berkeley Artificial Intelligence Research lab. 
% The principal focus of his PhD research was the development of approaches for increasing the efficiency of processes for providing supervision to guide autonomous systems with particular emphasis on transfer learning and learning from demonstration, work which was awarded as Best Student Paper at the  IEEE/RSJ International Conference on Intelligent Robots and Systems 2016.

% Being in the field of robotics since 2010, he has been previously part of research efforts on space exploration robots, GPU-based simulations and robotic platforms for first responders as well as mobile autonomy at leading research institutions including MIT, ETHZ and the University of Oxford.

\bibliographystyle{plain}
\bibliography{references}

\end{document}